\documentclass[journal]{IEEEtran}

\usepackage[noadjust]{cite}
\usepackage[colorlinks=true,citecolor=green,linkcolor=red,urlcolor=blue]{hyperref}
\usepackage{times}
\usepackage{epsfig}
\usepackage{graphicx}
\usepackage{amsmath}
\usepackage{amssymb}
\usepackage{verbatim}
\usepackage{booktabs} 
\usepackage{adjustbox}
\usepackage{footnote}
\usepackage{colortbl} 
\usepackage{xcolor} 
\usepackage{xfrac}
\usepackage[ruled,vlined,noresetcount]{algorithm2e}
\ifCLASSINFOpdf
\else
\fi
\begin{document}
%
\title{3D Convolutional Neural Networks for Cross Audio-Visual Matching Recognition}
%
%
%

\author{Amirsina~Torfi, Seyed Mehdi Iranmanesh,
        Nasser~Nasrabadi,~\IEEEmembership{Fellow,~IEEE}
        and~Jeremy~Dawson
        

}

\maketitle

\begin{abstract}
Audio-visual recognition~(AVR) has been considered as a solution for speech recognition tasks when the audio is corrupted, as well as a visual recognition method used for speaker verification in multi-speaker scenarios.~The approach of AVR systems is to leverage the extracted information from one modality to improve the recognition ability of the other modality by complementing the missing information. The essential problem is to find the correspondence between the audio and visual streams, which is the goal of this work.~We propose the use of a coupled 3D Convolutional Neural Network~(3D-CNN) architecture that can map both modalities into a representation space to evaluate the correspondence of audio-visual streams using the learned multimodal features.~The proposed architecture will incorporate both spatial and temporal information jointly to effectively find the correlation between temporal information for different modalities.~By using a relatively small network architecture and much smaller dataset for training, our proposed method surpasses the performance of the existing similar methods for audio-visual matching which use 3D CNNs for feature representation.~We also demonstrate that an effective pair selection method can significantly increase the performance. The proposed method achieves relative improvements over 20\% on the Equal Error Rate~(EER) and over 7\% on the Average Precision~(AP) in comparison to the state-of-the-art method.
\end{abstract}


%
\IEEEpeerreviewmaketitle

\section{Introduction}
%
%
%
%

\IEEEPARstart{T}{he} crucial part of an AVR algorithm is the feature selection for both audio and visual modalities, which has a direct impact on the performance of the audio-visual recognition task. Regarding the speech modality, most speech recognition systems employ Hidden Markov Models~(HMMs) to extract the temporal information of speech and Gaussian Mixture Models (GMMs) to discriminate between different HMMs states for acoustic input representation. However deep learning has recently been employed as a mechanism for unsupervised speech feature extraction~\cite{6296526}.~Beyond speaker and speech recognition, deep learning has also been used for feature extraction of unlabeled facial images~\cite{6639343}. Similar approaches have been employed in the analysis of multi-modal voice and face data, which resulted in an improvement of speech recognition performance~\cite{Ngiam2011MultimodalDL}.

The inference based on common sense is that the lip motions and the heard voice which is represented by speech features are highly correlated as a human is usually able to match the heard sound to a given set of lip motion. However, the visual lip motions and their corresponding audio stream still can have non-negligible uncorrelated information. Decision fusion has been shown to be effective in which the final decision is made by fusing the statistically independent decisions from different modalities with the emphasize on uncorrelated characteristics between different modalities \cite{wu2016novel}. However, data fusion in early stages, demonstrated more promising results as it creates a joint representation between two modalities based on the cross-modality correlations \cite{huang2013audio,ngiam2011multimodal}. As the corresponding audio-visual streams have correlated and uncorrelated information, we propose an architecture based on Deep Neural Networks~(DNNs) as a discriminative model between the two modalities in order to simultaneously distinguish between the correlated and uncorrelated components.

Alongside with the audio stream, lip motions can also contain speaker-related information. Some research efforts applied both modalities for Speaker Identification~(SI) and Speaker Verification~(SV) mainly based on decision fusion and MFCC features \cite{neti2000audio,erzin2005multimodal}. The speaker dependent systems are generally aimed to recognize the speech or speaker identity based on speaker-dependent characteristics. However, speaker-independent systems must be able to recognize the part of speech regardless of who is speaking. The SV has two general categories of text-dependent and text-independent types. In text-dependent setup, a fixed text is used for all experiments. On the other hand, in text-independent SV, no prior information or restrictions are considered for the utterances. It makes the text-independent to be more challenging that text-dependent scenario. Most of the previous research efforts for audio-visual for the aforementioned problems have been conducted in the text-dependent scenario. In contrast, we conduct our experiments in speaker-independent and text-independent mode to deliberately investigate the most challenging scenario.

There is a significant amount of literature describing audio-visual recognition in a variety of applications, including speech recognition in noisy environments using lip motions as auxiliary features \cite{zeiler2016robust,wang2016audio}, as well as the converse where speech data is leveraged for the purpose of lip reading \cite{almajai2016improved,werda2013lip}. However, there is a lack of research on concurrently incorporating the spatial and temporal audio-visual information to address the root problem of whether or not the audio stream and the visual stream match.~As an example, in a multi-speaker scenario, if the features connecting audio and video can be found,~the speakers' lip motions could be determined by the audio stream and vice versa. In this paper, we investigate the main problem of audio-visual matching. In another word, the main problem is to recognize whether the visual lip motions of a speaker corresponds to the accompanying speech signal. The aforementioned root problem is the precedent to audio-visual synchrony verification, as recognizing the consistency between the audio-visual streams is desired. The problem of audio-visual synchrony recognition has been addressed in different research efforts such as \cite{bredin2007audiovisual} for identity verification, and liveness recognition of the audio-visual streams \cite{slaney2001facesync,chetty2004liveness}.

To address the problem, we propose to use the 3D Convolutional Neural Networks models that have recently been employed for action recognition, scene understanding, and speaker verification and demonstrated promising results~\cite{ji20133d,tran2015learning,torfi2017text}.~3D CNNs concurrently extract features from both spatial and temporal dimensions, so the motion information is captured and concatenated in adjacent frames. We use 3D CNNs to generate separate channels of information from the input frames. The combination of all channels of correlated information creates the final feature representation.

The focus of the research effort described in this paper is to implement two non-identical 3D-CNNs for audio-visual matching~(Section~\ref{sec:architecture}). The goal is to design nonlinear mappings that learn a non-linear embedding space between the corresponding audio-video streams using a simple distance metric. This architecture can be learned by evaluating pairs of audio-video data and later used for distinguishing between pairs of matched and non-matched audio-visual streams.~One of the main advantages of our audio-visual model is the noise-robust audio features,~which are extracted from speech features with a locality characteristic~(Section~\ref{section:Data Representation}),~and the visual features, which are extracted from spatial and temporal information of lip motions. Both audio-visual features are extracted using 3D CNNs, allowing the temporal information to be treated separately for better decision making.

The contributions of this paper are as follows:

\begin{itemize}
\item A novel coupled 3D CNN architecture, which simultaneously extracts spatial and temporal information,~is designed with a significant reduction in dimension compared to the input space and is optimized for distinguishing between match and non-match audio-visual streams.

\item The network is relatively small, which has the advantage of allowing it to be easily trainable and fast to test.

\item Compared to traditional MFCCs, a different type of speech feature has been used for representing the audio stream, which provides more promising results.

\item An adaptive online pair selection method with the output feature space distance as a criterion has been proposed for selecting the main contributing pairs for accelerating the convergence speed and preventing over-fitting.

\end{itemize}

To the best of our knowledge, this is the first attempt to use 3D convolutional neural networks for audio-visual matching in which a bridge between spatio-temporal features has been established to build a common feature space between audio-visual modalities. Our source code\footnote{https://github.com/astorfi/lip-reading-deeplearning} has been released online as an open source project \cite{torfi3dconvaudiovisual}.

 
\section{Related works}

Lip reading and audio-visual speech recognition~(AVSR) are highly correlated such that the relevant information of one modality can improve the recognition of the other modality in any of the two aforementioned applications. DNNs have been employed for fusing speech and visual modalities for audio-visual automatic speech recognition~(AV-ASR)~\cite{7178347}.~Moreover, in~\cite{Noda:2015:ASR:2766528.2766557},~a connectionist HMM system is introduced for AVSR and a pre-trained CNN is used to classify phonemes.

Some researchers have used CNNs to predict and generate phonemes~\cite{a80e82ad3fd04b67b03238a113fd6a90} or visemes \cite{7406418} without considering the word-level prediction.~Phonemes are the smallest distinguishable unit of an audio stream which are combined to create a spoken word, and a viseme is its corresponding visual equivalent.~For recognizing full words,~a Long Short-Term Memory~(LSTM) classifier with Discrete Cosine Transform~(DCT) and Deep Bottleneck Features~(DBF) is trained~\cite{7472088}.~Similarly,~LSTM with Histogram of Oriented Gradients~(HOG) features are used also for phrase recognition~\cite{7472852}.

 One of the most challenging applications of audio-visual recognition is the audio-video synchronization for which the audio-visual matching ability is required.~The research efforts related to this paper consist of different audio-visual recognition tasks which try to find the correspondence between the two modalities.~Different approaches have been employed for tackling the audio-visual matching problem. Some are based on data-driven approaches, such as using DNN classifiers to determine the off-sync time~\cite{Chung16a,Marcheret2015DetectingAS} and some are based on Canonical Correlation Analysis~(CCA)~\cite{sargin2007audiovisual} and Co-Inertia Analysis~(CoIA).

 The most relevant work to ours is the work of Chung and Zisserman~\cite{Chung16a} which is aimed at determining the audio-video synchronization between lip motions and an audio stream in a video.~They use traditional Mel-frequency cepstral coefficients~(MFCCs) to present speech features.~In~\cite{Chung16a}, CNNs have been applied for feature construction with the motion information as the input depth, which does not effectively reflect the correlation and distinction between spatial and temporal information. Instead,~we propose to apply 3D convolutional layers to simultaneously capture the spatial and temporal discriminative features independently. Moreover, we designed an adaptive pair selection method for improving the accuracy and accelerating the training convergence, as opposed to using the whole training data which has been used in~\cite{Chung16a}.

\section{Dataset}\label{Dataset}

The datasets that have been used for our experiments are the \emph{Lip Reading in the Wild (LRW)} \cite{Chung16} and the \emph{WVU Audio-Visual Dataset} dataset~(\emph{AVD})~\cite{wvudataset}.~The LRW dataset consists of up to 1000 utterances of 500 different words, spoken by different speakers.~All videos are 1.16 seconds in length, and the word occurs in the middle of the video.~The \emph{AVD} dataset consists of audio and video data collected over a period spanning 2014 through 2015.~The video and audio data consists of both scripted and unscripted voice samples.~For the scripted samples, the participant read a sample of text.~For the unscripted samples,~the participant answered interview questions that prompted conversational responses rather than simple `yes' or `no' answers.

\subsection{Processing}

\begin{figure}[!t]
\centering
\includegraphics[width=0.8\linewidth]{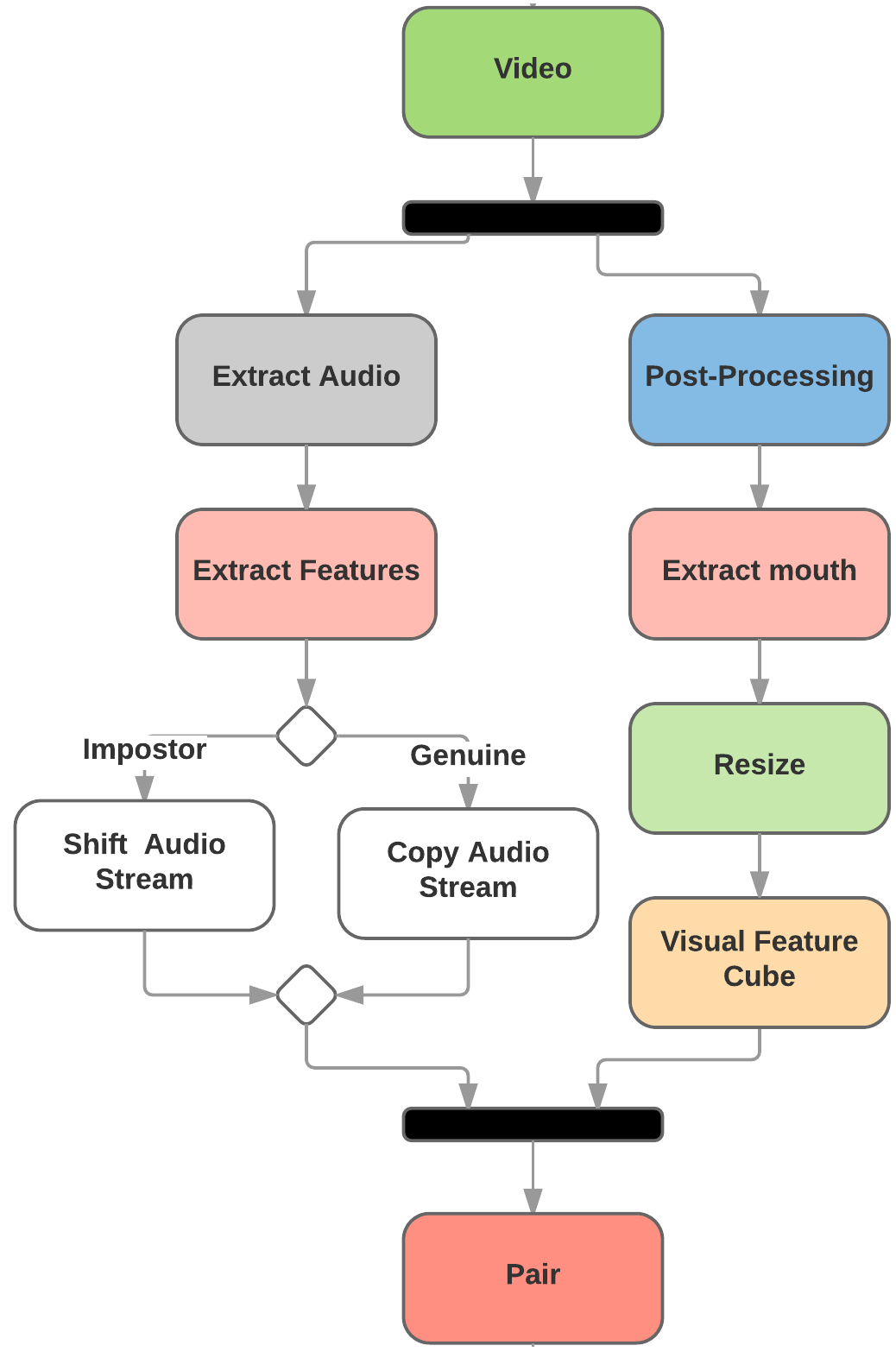}
\caption{The processing pipeline of the datasets.}
   
\label{fig:processing-pipeline}
\end{figure}

The processing pipeline of both datasets is shown in Fig.~\ref{fig:processing-pipeline}. The pipeline is subdivided into two visual and audio sections.~In the visual section, the videos are post-processed to have equal frame rate of 30 f/s.~Then,~face tracking and mouth area extraction is performed on the videos using the dlib library~\cite{dlib09}.~Finally, all mouth areas are resized to have the same size, and concatenated to form the input feature cube.~The dataset does not contain any audio files. In the audio section, the audio files are extracted from videos using the FFmpeg framework~\cite{FFmpeg}. Then the speech features will be extracted from audio files using SpeechPy package~\cite{torfispeechpy}.

%
%
%

\section{Data representation}\label{section:Data Representation}

The proposed architecture utilizes two non-identical ConvNets which uses a pair of speech and video streams.~The network input is a pair of features that represent lip movement and speech features extracted from 0.3-second of a video clip.~The main task is to determine if a stream of audio corresponds with a lip motion clip within the desired stream duration. 

The difficulty of this task is the short time interval of the video clip~(0.3-0.5 second) considered to evaluate the method.~This setting is close to real-world scenarios because, in some biometrics or forensics applications, only a short amount of captured video or audio might be available to distinguish between different modalities.~Temporal video and audio features must correspond over the time interval they cover.~This correspondence is discussed in the next two sections.

\subsection{Speech}

The main characteristic of CNNs is their locality, i.e., the convolution operation is applied to specific local regions in an image. As a visual inference of this locality property, the neighbor features should be correlated in some sense.~Since the input speech feature maps are treated as images when a CNN architecture is used, the features must be locally correlated in the sense of time and frequency on both axes respectively. 

The MFCCs, which are derived from the cepstral representation of the audio stream, can be used as the speech feature representation.~However,~the drawback is that they have non-local characteristics.~The reason for this is that during the last operation~(DCT\footnote{Discrete Cosine Transform}) for generating MFCCs, which is aimed at eliminating the correlations between energy coefficients, the order of the filter-bank energies is changed, which leads to disturbing the locality property.~The approach employed in this paper is to use the log-energies derived directly from the filter-banks energies which we call MFECs\footnote{Mel-frequency energy coefficients}.~The extraction of MFECs is similar to MFCCs, but with no DCT operation.

 \begin{figure}[!t]
\begin{center}
   \includegraphics[width=1.0\linewidth]{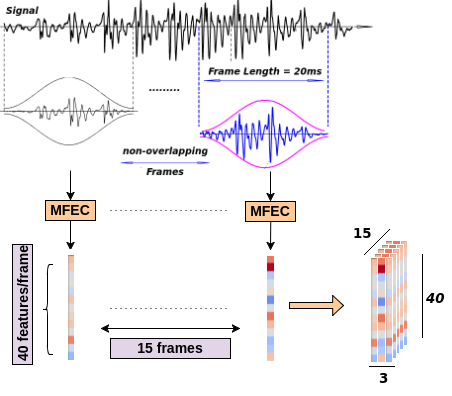}
\end{center}
   \caption{Speech feature generation using stacked frames created from the input signal samples.}
\label{fig:speech_feature_map}
\end{figure}

On the time axis, the temporal features are non-overlapping 20ms windows which are used for the generation of spectrum features that possess a local characteristic.~The input speech feature map, which is represented as an image cube, corresponds to the spectrogram, as well as the first and second order derivatives of the MFEC features.~These three channels correspond to the image depth.~This representation is depicted in Fig.~\ref{fig:speech_feature_map}.~Collectively, from a 0.3-second clip, 15 temporal feature sets~(each forms a 40 MFEC features) can be derived which form a speech feature cube.~Each input feature map for a single audio stream has a dimensionality of $15\times40\times3$.

%

Similar approaches are used in \cite{6857341} for automatic speech recognition~(ASR), where local filtering layers are employed to extract and represent the spatial speaker-independent features.~Similar input features are employed in~\cite{Chung16a}.~However, these efforts used MFCC features as the speech feature representations without the use of the temporal derivatives.

\subsection{Video}

\begin{figure}[t]
\begin{center}
   \includegraphics[width=1\linewidth]{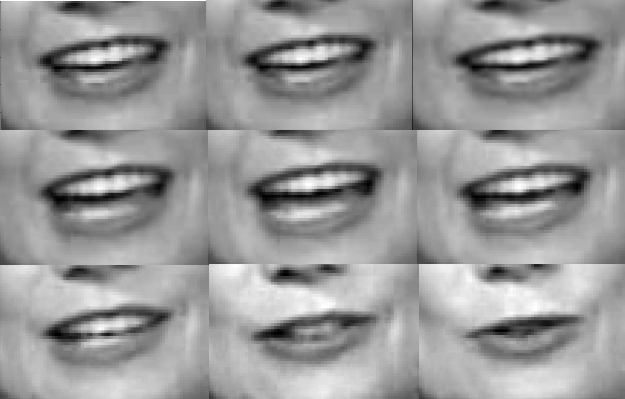}
\end{center}
   \caption{The sequence of mouth areas in a 0.3-second video stream.}
\label{fig:video_frame}
\end{figure}

The frame rate of each video clip used in this effort is 30 f/s.~Consequently, 9 successive image frames form the 0.3-second visual stream.~The input of the visual stream of the network is a cube of size 9x60x100,~where 9 is the number of frames that represent the temporal information.~Each channel is a 60x100 gray-scale image of mouth region. An example of mouth area representation is provided in Fig. \ref{fig:video_frame}.

A relatively small mouth crop region is intentionally chosen due to practical consideration because, in real-world scenarios, it is not very likely to have access to high-resolution images.~Moreover,~unlike the usual experimental setups for CNNs, we did not restrict our experiments to input images with uniformly square aspect ratios.

\subsection{Preprocessing}

Neither mean-feature-normalization\footnote{$\frac{X-\bar{X}}{X_{max}-X_{min}}$} nor Min-Max scaling\footnote{$\frac{X-X_{min}}{X_{max}-X_{min}}$} demonstrated promising results in the testing phase for our experiments.~However,~both improved the training optimization.~We found the data standardization ($\frac{X-\bar{X}}{\sigma}$) to be effective as a preprocessing operation.

\section{Architecture}\label{sec:architecture}

\begin{figure}[t]
\begin{center}
   \includegraphics[width=1\linewidth]{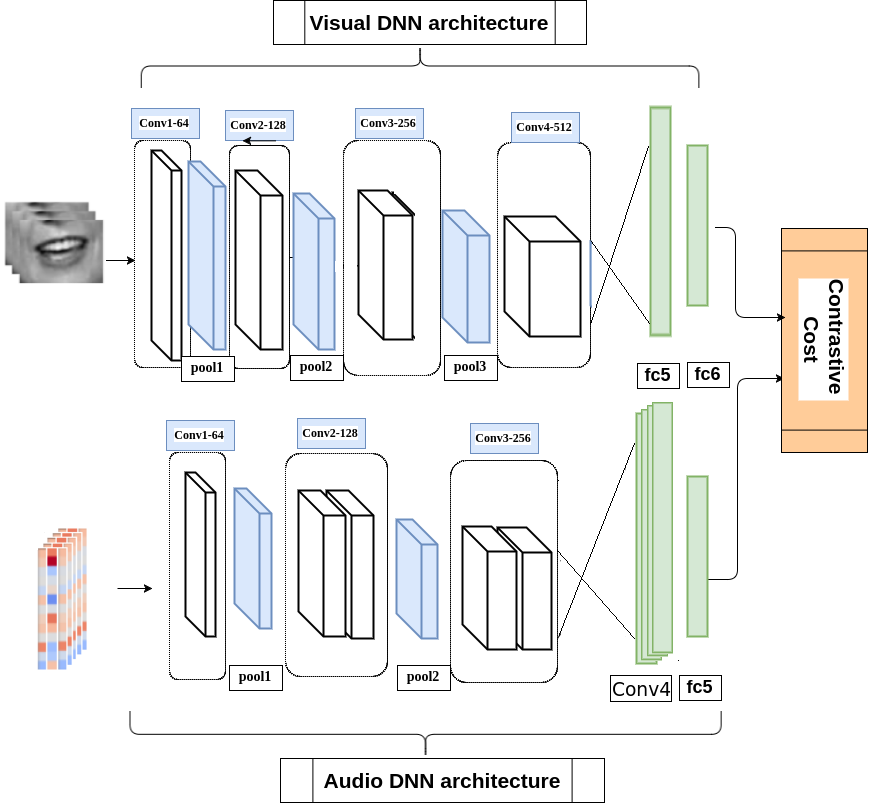}
\end{center}
   \caption{\textbf{Outline of the architecture.} Two non-identical 3D CNNs with audio and visual streams as inputs are coupled together in the last fully-connected layer by contrastive cost criterion.}
\label{fig:architecture}
\end{figure}

The architecture is a coupled 3D convolutional neural network in which two different networks with different sets of weights must be trained.~For the visual network,~the lip motions' spatial and temporal information are incorporated jointly and will be fused for exploiting the temporal correlation.~For the audio network, the extracted energy features are considered as a spatial dimension, and the stacked audio frames create the temporal dimension.~In our proposed 3D CNN architecture, the convolutional operations are performed on successive temporal frames for both audio-visual streams.~Dropout($\rho$) has been used for all fully-connected layers prior to the last layer.~Except for the last layer, all layers are followed by PReLU activation, as proposed in~\cite{He:2015:DDR:2919332.2919814} which is a generalization to ReLU.~Compared to ReLU activation, PReLU employment demonstrated better performance in our experiments in which the parameters for rectifiers are learned adaptively.~The architecture is depicted in Fig.~\ref{fig:architecture}.

\subsection{Visual network}

The network architecture used for training on video streams is described in Table~\ref{table:mouth-cnn}.~In Table~\ref{table:mouth-cnn},~the spatial size of the 3D kernels is reported as $T \times H \times W$ where $T$ is the kernel size in temporal dimension, and $H$ and $W$ are the kernel sizes in height and width dimensions,~respectively.~As with usual CNN architectures, the kernel depth is equal to the input-channel size, which is the number of the feature maps of the previous layer.~For kernel spatial size representation, the kernel depth sizes are discarded for simplicity.


\begin{table}[h]
\caption[Table caption text]{The architecture for video stream.}
\label{table:mouth-cnn}
\begin{center}
\begin{tabular}{ccccc}
\toprule 
layer & input-size & output-size & kernel & stride \\
\hline
\midrule
\rowcolor{black!10} Conv1 & 9x60x100x1 & 7x58x98x16 & 3x3x3 & 1 \\
Pool1 & 7x58x98x16 & 7x28x48x16 & 1x3x3 & 1x2x2 \\
\rowcolor{black!10} Conv2 &  7x28x48x16 & 5x26x46x32 & 3x3x3 & 1 \\
Pool2 & 5x26x46x32 & 5x12x22x32 & 1x3x3 & 1x2x2 \\
\rowcolor{black!10} Conv3 & 5x12x22x32 & 3x10x20x64 & 3x3x3 & 1 \\
Pool3 & 3x10x20x64 & 3x4x9x64 & 1x3x3 & 1x2x2 \\
\rowcolor{black!10} Conv4 & 3x4x9x64 & 1x2x7x128 & 3x3x3 & 1 \\

\rowcolor{black!20}FC5 & 1x2x7x128 & 256 & - & - \\
\rowcolor{black!20}FC6 & 256 & 64 & - & - \\
\bottomrule
\end{tabular}
\end{center}

\end{table}

An important characteristic of the visual network is its pooling method.~Since we are using 3D convolutional layers,~3D pooling layers are also utilized.~Although 1x3x3 kernels are applied for spatial feature pooling, in order to increase robustness to the moving lip effect\footnote{The lip place in the video clip is not necessarily fixed},~the pooling stride is set to two\footnote{It means a kind of overlap because the pooling kernel is 1x3x3} in order to maintain lip movement features\footnote{More importantly high-level features} in the neighborhood of the pooling kernel.~The 3D convolutional operations are performed to find the correlation between high-level temporal and spatial information by fusion among them.~No zero-padding is used in the visual architecture.

\subsection{Audio network}

The network architecture used for training on audio streams is described in Table~\ref{table:speech-cnn}.~In our architecture, pooling operations are only applied in the frequency axis~(domain) to maintain the temporal information within the time frames.~Additionally, our proposed architecture has a high level of compression in which only 64 output units are used.

As with Table~\ref{table:mouth-cnn},~in Table~\ref{table:speech-cnn},~the spatial size of the 3D kernels is reported as $T \times H \times W$, and ~the kernel depth sizes are discarded for simplicity as well.~A 3D kernel is used in the first layer for spatio-temporal feature extraction using a 3D convolutional operation.~Except for the first layer, since the spatial dimension for the audio feature maps are $M \times N \times 1$ in higher-level layers, we are, in essence, dealing with 2D dimensionality.~Because of this, we have regular 2D convolutional operations for the audio network which simultaneously capture the temporal and spatial information using their 2D kernels.~The kernel spatial dimensions are deliberately demonstrated as $T \times H \times 1$ to emphasize the connection between the 3D and 2D convolutional operations.

\begin{table}[h]
\caption[Table caption text]{The architecture for audio stream.}
\label{table:speech-cnn}
\begin{center}
\begin{tabular}{ccccc}
\toprule 
layer & input-size & output-size & kernel & stride \\
\hline
\midrule
\rowcolor{black!10} Conv1 & 15x40x3 & 13x36x1x16 & 3x5x3 & 1 \\
Pool1 & 13x36x1x16 & 13x18x1x16 & 1x2x1 & 1x2x1 \\
\rowcolor{black!10}Conv2-1 & 13x18x1x16 & 11x15x1x32 & 3x4x1 & 1 \\
\rowcolor{black!10}Conv2-2 & 11x15x1x32 & 9x12x1x32 & 3x4x1 & 1 \\
Pool2 & 9x12x1x32 & 9x6x1x32 & 1x2x1 & 1x2x1 \\
\rowcolor{black!10}Conv3-1 & 9x6x1x32 & 7x4x1x64 & 3x3x1 & 1 \\
\rowcolor{black!10}Conv3-2 & 7x4x1x64 & 5x2x1x64 & 3x3x1 & 1 \\

\rowcolor{black!15}Conv4 & 5x2x1x64 & 3x1x1x128 & 3x2x1 & 1 \\
\rowcolor{black!20}FC5 & 3x1x1x128 & 64 & - & - \\
\bottomrule
\end{tabular}
\end{center}

\end{table}

In the audio architecture,~zero-padding is not used because zero-padding adds extra virtual zero-energy coefficients which are meaningless in the sense of local feature extraction.~Another important characteristic is the use of non-square kernels.~As we go from low-level features~(lower level convolutional layers) to high-level features~(higher level convolutional layers), the kernel widths follow a decreasing order.~This setup results in extraction and processing of more temporal features in the lower level that are related to speech features, as well as and correlated features in the high-level features that are the features extracted from the CNN.

\subsection{Coupling}

Both audio-visual networks are coupled at their highest level, which is the last fully-connected layer with a cardinality of $\zeta$.~The $\zeta$ parameter represents the cardinality of the output embedding.~The default value for $\zeta$ is 64, which forms our base experiments.~Since the network is trained in verification mode, in order to match the audio and visual representation of the spoken parts of speech, a contrastive loss is used as a discriminative distance metric to optimize the coupling process, which has been proposed in \cite{Chopra:2005:LSM:1068507.1068961}. The contrastive loss function $L_W(Y,X)$ is as follows:

\begin{align}\label{eq2}
L_W(Y,X) = {{1}\over{N}}  \sum_{i=1}^{N} L_W(Y_i,(X_{p_{1}},X_{p_{2}})_i),
\end{align}

\noindent where N is the number of training samples, $(X_{p_{1}},X_{p_{2}})_i$ is the i-th input pair, $Y_i$ is the corresponding label and $L_W(Y_i,(X_{p_{1}},X_{p_{2}})_i)$ is defined as follows:

\begin{equation} \label{eq3}
\begin{split}
L_W&(Y_i,(X_{p_{1}},X_{p_{2}})_i) = Y \times L_{gen}(D_W(X_{p_{1}},X_{p_{2}})_i)\\ &+ (1-Y) \times L_{imp}(D_W(X_{p_{1}},X_{p_{2}})_i)+\lambda{||W||}_{2},
\end{split}
\end{equation}

\noindent in which $D_W(X_{p_{1}},X_{p_{2}})$ is the Euclidean distance between the outputs of the network with $(X_{p_{1}},X_{p_{2}})$ as the input.~The last term is for regularization, and $\lambda$ is the regularization parameter.~Finally, $L_{gen}$ and $L_{imp}$ are defined as the functions of $D_W(X_{p_{1}},X_{p_{2}})$ by the following equations:

\begin{equation}\label{eq4}
  \begin{cases}
    L_{gen}(D_W(X_{p_{1}},X_{p_{2}}))={{1}\over{2}}{D_W(X_{p_{1}},X_{p_{2}})}^2\\
    L_{imp}(D_W(X_{p_{1}},X_{p_{2}}))={{1}\over{2}}max\{0,{(\mu-D_W(X_{p_{1}},X_{p_{2}}))}\}^2,
  \end{cases}
\end{equation}

\noindent where $\mu$ is the predefined margin.~Contrastive loss is employed as a mapping criteria, which will ideally place genuine pairs\footnote{Also called as match pair in which both samples in a pair belong to the same identity} to nearby and impostor pairs\footnote{Also called as non-match pair in which samples in a pair belong to the different identity} to distant manifolds in the output space. 

\section{Evaluation and verification metric}\label{sec:evaluation}
In this paper, we evaluate experimental results using the Receiver Operating Characteristic~(ROC) and Precision-Recall~(PR) curves characteristics.~The ROC curve consists of the Validation Rate~(VR) and False Acceptance Rate~(FAR).~All pairs $(X_{P_1},X_{P_2})$ of the same identity are denoted with $\mathcal{P}_{gen}$ whereas all pairs belonging to different identities are denoted as $\mathcal{P}_{imp}$.~If $D_W$ is the Euclidean distance between the outputs of the network with $(X_{P_1},X_{P_2})$ as the input, we define true positive and false acceptance as below:

\begin{equation}\label{eq5}
  \begin{split}
    TP(\tau) = \left\{ (X_{P_1},X_{P_2})\in \mathcal{P}_{gen}, D_W\leq \tau \right\}.
  \end{split}
\end{equation}
\begin{equation}\label{eq6}
  \begin{split}
    FA(\tau) = \left\{ (X_{P_1},X_{P_2})\in \mathcal{P}_{imp}, D_W\leq \tau \right\}.
  \end{split}
\end{equation}

Here, $TP(\tau)$ is the test samples which are classified as match pairs, whereas $FA(\tau)$ are non-match pairs which are incorrectly classified as match pairs. Both calculations are done using a single pre-defined threshold which the output distance will be compared with it as a metric for prediction. 

True Positive Rate~(TPR) and the False acceptance rate~(FAR) are calculated as below:

\begin{equation}\label{eq7}
  \begin{split}
    TPR(\tau)=\frac{TP(\tau)}{\mathcal{P}_{gen}},FAR(\tau)=\frac{FA(\tau)}{\mathcal{P}_{imp}}.
  \end{split}
\end{equation}

The TPR is the percentage of genuine pairs that are correctly classified as match pairs and it is called \emph{Recall}. On the other hand, the FAR is the percentage of non-match pairs (impostor pairs) that are incorrectly classified as match pairs. According to the definitions, FAR and TPR can be computed with regard to impostor and genuine pairs, respectively. 

 Another metric which is called \emph{Precision}, is widely used for accuracy evaluation.~It is defined as below:

\begin{equation}\label{eq7}
  \begin{split}
    Precision(\tau)=\frac{TP(\tau)}{FA(\tau) + TP(\tau)}.
  \end{split}
\end{equation}

In comparison to Recall, Precision is the portion of retrieved positive-classified samples that are correctly classified, while recall is the portion of positive samples that are retrieved as positive. Basically, Precision and Recall are inversely correlated. The precision-recall (PR) curve demonstrates the balance between two aforementioned metrics. 

Precision-Recall (PR) curves are often used in Information Retrieval and considered as an alternative to ROC curves for classification tasks with a large difference in the class distribution \cite{Davis:2006:RPR:1143844.1143874}. 

The main metric that has been used for performance evaluation is the Equal Error Rate~(EER) which is the point when FAR and FRR are equal.~Moreover, Area Under the Curve~(AUC) is evaluated as the accuracy, which is the area under the ROC curve.~Average-Precision~(AP) is another employed metric which corresponds to the area under the Precision-Recall~(PR) curve. Since we do not restrict our experiments to have the same or even close portion of genuine and impostor pairs, the AP metric can be more reliable as a representative of the accuracy which belongs to the PR curve. 

For verification, the metric is simply a $\ell_2-norm$ calculation between the outputs of the two fully connected layers from the two parallel CNNs, and a final comparison with a given threshold. To provide a better statistical demonstration of the performance,~the test samples were split into 5 disjoint parts,~and the averaged performance is reported across the five splits. In essence, 5-fold validation has been employed for evaluation. All performance evaluations are reported based on the statistics of the result as $\mu \pm \sigma$ using the 5-fold validation.

\section{Training}\label{Training}

In this section,~the method and manner of training will be described.~A variance scaling initializer that has been recently developed for network weight initialization \cite{He:2015:DDR:2919332.2919814} is used for our experiments.~Batch normalization \cite{Ioffe2015BatchNA} has been used for improving the training convergence. The batch size is $32$ for all of the experiments.


\subsection{Online pair selection}\label{Online Pair Selection}

The pair selection in our experiments is similar to the one used in \cite{Schroff_2015_CVPR}, in the sense of choosing challenging pairs.~However,~there are differences.~First, and most importantly, unlike the face verification, in our experiments~(AV matching), anchors cannot be defined directly as the representatives of the classes. The reason is because the pairs are defined as a one-vs-one in an audio-visual stream, and cannot be defined as a global class as can be done in image verification application. Second,~since there is no anchor,~the triplet loss cannot be defined as the one in \cite{Schroff_2015_CVPR},~so we propose a method for adaptive-thresholding which will be described later in this section.

All genuine pairs are selected, and the pair selection process is restricted to only choosing main contributing impostor pairs.~The criterion for choosing pairs is the distance between them in the output embedding feature space.~Assume the embedding feature space is represented by $f(x) \in R^{D}$.~If the input vector is $x$, the output is $f(x)$ with dimensionality of $D$.~The no-pair-selection case is when all the impostor pairs ($X^{imp}_{p}$) lead to larger output distances than all the genuine pairs~($X^{gen}_{p}$).~This leads to:

\begin{align}\label{eq:pairselection} 
\left \| X^{imp}_{p_{i}} - X^{gen}_{p_{j}} \right \| > \eta , \forall (X^{imp}_{p_{i}},X^{gen}_{p_{j}}) \in Mini-batch,
\end{align}

\noindent in which $\eta$ is an adaptive threshold which set a margin between impostor and genuine pairs. Algorithm~\ref{algorithm:pair selection} demonstrates the procedure.

\begin{algorithm}\label{algorithm:pair selection} 
 \KwData{extract mini-batch;}
 \textbf{Update}: Do not run optimizer~(no weight update)
 initialization\;
 \textbf{Evaluation}: Feed pairs and generate output distance vector\;
 \textbf{Find}: Find the maximum and minimum distances belonging to genuine pairs : $max\_gen$ \& $min\_gen$\;
 \textbf{Adaptive Threshold}: Calculate $\eta=\eta_{0}\times \left | \frac{max\_gen}{min\_gen}\right|$\;
 \While{checking impostor pairs}{
  evaluate the current impostor pair output distance: $imp\_dis$\;
  \eIf{$imp\_dis > max\_gen + \eta$}{
   discard current impostor pair\;
   }{
   select current impostor pair and return its index\;
  }
 }
 \caption{The adaptive online pair selection algorithm for selecting the main contributing impostor pairs in each mini-batch.}
\end{algorithm}

As it can be observed from Algorithm~\ref{algorithm:pair selection}, each mini-batch has its own threshold which will adaptively follow a descending order as the genuine pairs become closer on the output manifold space. Empirically we found this method to accelerate the convergence speed and moreover improved the accuracy.

\subsection{Hyperparameter optimization}

The hyperparameters in our experiments are: $\mu$ the margin for the contrastive loss, $\lambda$ the regularization parameter, dropout parameter~($\rho$) and the $\eta_{0}$ as the initial margin between impostor and genuine pairs in the pair selection phase.~The k-fold cross-validation method is used to estimate the hyperparameters in which we set $k=5$. 

In \emph{k-fold} cross-validation, the original training data is randomly divided into $k$ equal parts. Of the k-parts, one of them is fixed as the validation data for testing the model, and the other $k-1$ parts are used as training data. The cross-validation process is then repeated 5 times and the average error is used to determine the best parameter.

The online pair selection is not used in the cross-validation phase. It is worth noting that the data splitting is done per subject and not just the randomly selected data. Essentially, the data is split into $k=5$ equal parts such that none of the subjects present in one part are available in any other part. This has the practical advantage of preventing subject-specific characteristics from affecting the accuracy on the test split.

\section{Experiments and results}\label{sec:Experiments}

The experiments of this section have been conducted on the audio-visual matching task to evaluate the effectiveness of the employed architecture.~In evaluation of the experiments, we use the setup described in Section~\ref{Training}. 

\subsection{Evaluation on LRW dataset}\label{Oxford Dataset Experiments}
For audio-visual matching using the \emph{Lip Reading in the Wild} dataset, 500 words~(subjects) are available.~To make the train and test sets mutually exclusive, the first 400 words are used for creating the training set and the remaining 100 words are used for test set generation.~For each of the train/test sets,~only 50 utterances of each word are chosen for data generation.~The compiled initial training data contains generated genuine and impostor pairs.~The reason this is described as initial training data is that not all the generated data is used for training.~The method of selecting pairs was described in Section \ref{Online Pair Selection}.~The train/test characteristics are summarized in Table~\ref{table:oxford-dataset-summary}. 
\begin{table}[h]
\caption[Table caption text]{The train/test data summary for Lip Reading in the Wild dataset.}
\label{table:oxford-dataset-summary}
\begin{center}
\begin{tabular}{cccc}
\toprule 
set & \# subjects(words) & \# word utterances & \# pairs \\
\hline
\midrule
\rowcolor{black!10} train & 400 & 50 & 280k \\
test & 100 & 50 & 70k \\
\bottomrule
\end{tabular}
\end{center}

\end{table}

\begin{figure}[t]
\begin{center}
   \includegraphics[width=1.0\linewidth]{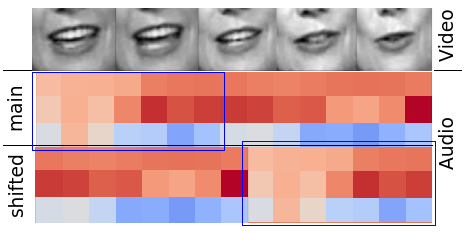}
\end{center}
   \caption{Audio and video feature maps for pair generation.~The shifted part is shown by the blue rectangular as an example.}
\label{fig:pair-generation}
\end{figure}

Genuine pairs~(audio/video) are created by matching the 9-channel visual feature cube with the corresponding audio feature cube as we discussed earlier in Section~\ref{section:Data Representation}.~For impostor pair generation, the audio feature map for a video is shifted alongside its time axis.~The shifting is random, and could be up to 0.5-second at maximum.~This shifting method allows the network to learn the matching between audio-video streams.~The pair generation method is depicted in Fig.~\ref{fig:pair-generation}.

Different experiments have been conducted to investigate the effects of the architecture,~feature selection,~and pair selection method.~In all experiments performed to generate test data,~unless otherwise stated, we used a 0.5-second shift for generating impostor pairs,~experiments were performed using MFEC features~(by using first and second order derivatives as well), and the output embedded feature space dimensionality was chosen to be 64.~The training stops after 15 epochs of training data,~or if the test accuracy shows descending behavior,~whichever occurs first.

To demonstrate the performance of our experimental results,~we report here,~in tabular form,~the EER and AUC values extracted from ROC curves and the AP value extracted from Precision-Recall curves. \\

\noindent \textbf{Effect of the proposed architecture and data representation}\\

In this section,~the effect of choosing different data representation~(MFCC/MFEC features \& using temporal derivatives) on the performance results is investigated.~We compare our method with the state-of-the-art in \cite{Chung16a} in which regular CNN architecture with MFCCs for speech feature representation are used. In \cite{Chung16a} only one channel MFCC has been used. We modify the structure represented in \cite{Chung16a} to accept 3-channels input speech features as we use MFCC alongside with its first and second order derivatives. 

For having a more comprehensive comparison, we compare our method with two common audio-visual synchrony approaches, based on CCA \cite{sargin2007audiovisual} and CoIA \cite{rua2009audio}. Prior to CCA/CoIA transformation, since the audio features are extracted at a faster rate, we interpolated the visual lip motions to have the same frame rate as audio features. Empirical evidence showed that not all the canonical correlations carry useful information. Considering the aforementioned evidence, only 20 dimensions of the correlation feature vector extracted from CCA or CoIA operations on audio-visual features are chosen which are corresponding to the higher correlation coefficients. For speech feature representation, in addition to static MFCC features, first and second order derivatives have been used as well.

 The results are summarized in Table~\ref{table:effect-of-feature}.~The \emph{depth} is the number of input channels, which is three if the first and second feature derivatives of MFEC/MFCC are used alongside the main features.

\begin{table}[h]
\caption[Table caption text]{Comparison of different methods on LRW dataset.}
\label{table:effect-of-feature}
\begin{center}
\addtolength{\tabcolsep}{-4pt}
\begin{tabular}{ccccc}
\toprule 
architecture & feature-depth & EER & AUC & AP \\
\hline
\midrule
\rowcolor{black!0} CCA\cite{sargin2007audiovisual} & MFCC-3 & 21.3\% $\pm$ 0.88 & 85.3\% $\pm$ 0.63 & 84.8\% $\pm$ 0.69 \\
\rowcolor{black!0} CoIA\cite{rua2009audio} & MFCC-3 & 20.6\% $\pm$ 0.79 & 85.8\% $\pm$ 0.81 & 86.0\% $\pm$ 0.59 \\
\rowcolor{black!5} CNN\cite{Chung16a} & MFCC-1 & 17.3\% $\pm$ 0.91 & 90.8\% $\pm$ 0.74 & 90.2\% $\pm$ 0.53 \\
\rowcolor{black!5} CNN\cite{Chung16a} & MFCC-3 & 17.6\% $\pm$ 1.12 & 91.0\% $\pm$ 1.22 & 90.4\% $\pm$ 1.31 \\
\rowcolor{black!10} 3D-CNN & MFEC-1 & 15.1\% $\pm$ 0.52 & 94.2\% $\pm$ 0.37 & 95.1\% $\pm$ 0.61 \\
\rowcolor{black!10} \textbf{3D-CNN} & \textbf{MFEC-3} & \textbf{13.5\% $\pm$ 0.71} & \textbf{95.4\% $\pm$ 0.36} & \textbf{96.5\% $\pm$ 0.44} \\
\bottomrule
\end{tabular}
\end{center}

\end{table}

In the case of using MFCC features,~using first- and second-order derivatives did not improve the performance, and additionally, using a 3-channel input increased the variance of the performance.~This means that the stability of the results decreased.~As can be observed in Table~\ref{table:effect-of-feature}, using the MFEC feature alongside the temporal derivatives achieves the best result.\\

\noindent \textbf{Effect of embedding layer}\\

As mentioned earlier, the default cardinality~($\zeta$) for the embedding layer is 64.~In this section, we change the dimensionality of the embedding layer in order to evaluate its effect on performance, specifically, to observe the effect of feature compression.~The results are shown in Table~\ref{table:effect-of-embedding}.

\begin{table}[h]
\caption[Table caption text]{The effect of embedding layer on performance.}
\label{table:effect-of-embedding}
\begin{center}
\addtolength{\tabcolsep}{1pt}
\begin{tabular}{cccc}
\toprule 
dim & EER & AUC & AP \\
\hline
\midrule
\rowcolor{black!10} 16 & 14.0\% $\pm$ 0.83 & 95.0\% $\pm$ 0.22 & 96.0\% $\pm$ 0.51 \\
32 & 13.8\% $\pm$ 0.81 & 95.2\% $\pm$ 0.42 & 96.2\% $\pm$ 0.57 \\
\rowcolor{black!10}\textbf{64} & \textbf{13.5\% $\pm$ 0.71} & \textbf{95.4\% $\pm$ 0.36} & \textbf{96.5\% $\pm$ 0.44} \\
128 & 13.6\% $\pm$ 0.61 & 95.6\% $\pm$ 0.54 & 96.3\% $\pm$ 0.62 \\
\rowcolor{black!10}256 & 14.1\% $\pm$ 0.91 & 94.9\% $\pm$ 0.42 & 96.1\% $\pm$ 0.48 \\
\bottomrule
\end{tabular}
\end{center}

\end{table}

The results summarized in Table~\ref{table:effect-of-embedding} indicate that the performance changes due to variation in embedding dimensionality are not significant.~However,~at the highest dimensionality, a decrease can be seen.~This is due to overfitting caused by using too many representative features.\\

\noindent \textbf{Effect of online pair selection}\\

\begin{figure}[t]
\begin{center}
   \includegraphics[width=1.1\linewidth]{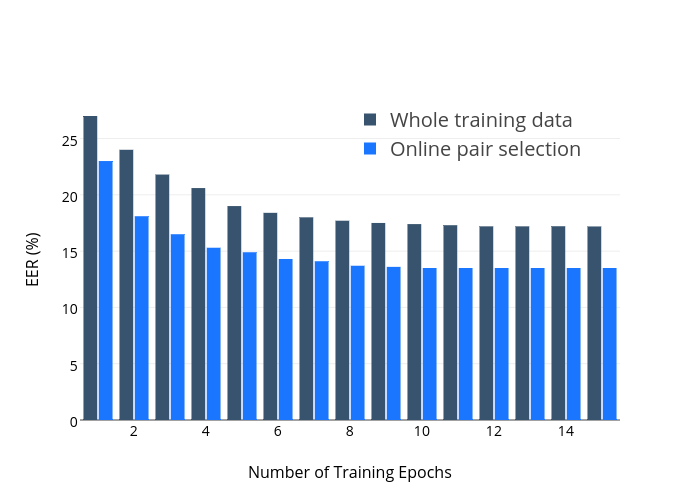}
\end{center}
   \caption{The effect of the proposed adaptive online pair selection method on the speed of convergence and matching ability.}
\label{fig:convergence-speed}
\end{figure}

\begin{figure}[t]
\begin{center}
   \includegraphics[width=1.1\linewidth]{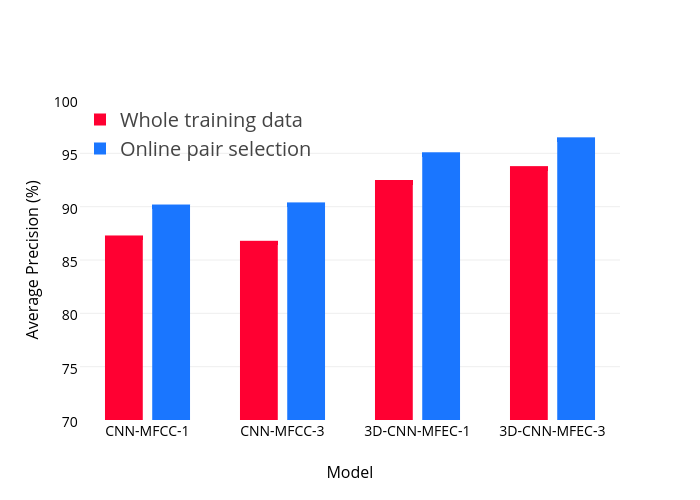}
\end{center}
   \caption{The comparison of proposed adaptive online pair selection method and choosing whole training data for different architectures.}
\label{fig:pairselection-ROC}
\end{figure}

The method for generating pairs has been described in Section \ref{Online Pair Selection}.~Here, we demonstrate the results with and without online pair selection.~The setup is the default described earlier, which is using 3-channels MFEC features with the embedding dimensionality of 64.~In addition to increasing the accuracy, the online pair selection resulted in a faster convergence in the training loss. Moreover, it was also faster in achieving the highest test accuracy.~The EER is reported for different epochs of training for both setups.~The averaged results for 5-runs of training are shown in Fig.~\ref{fig:convergence-speed},~concurrently illustrating the improvement in performance and speed of convergence in which the EER belongs to evaluation on test data per number of training epochs. The effect of online pair selection on the accuracy, is illustrated in Fig.\ref{fig:pairselection-ROC} for the default setup.\\

\noindent \textbf{The Effect of Time Shift}\\

In this section, the effect of time shift on the performance will be investigated.~However, here we do not make any changes in the training set. In this setup, we choose different shifts for generating impostor pairs solely in the test set in order to demonstrate the difficulty of the task.~The results are shown in Table~\ref{table:effect-of-shift}.

\begin{table}[h]
\caption[Table caption text]{The Effect of Shift for generating impostor pairs in the test set}
\label{table:effect-of-shift}
\begin{center}
\addtolength{\tabcolsep}{1pt}
\begin{tabular}{cccc}
\toprule 
shift(sec) & EER & AUC & AP \\
\hline
\midrule
0.3 & 17.3\% $\pm$ 0.74 & 90.0\% $\pm$ 0.68 & 89.4\% $\pm$ 0.53 \\
\rowcolor{black!10}0.4 & 14.9\% $\pm$ 0.82 & 94.0\% $\pm$ 0.45 & 94.1\% $\pm$ 0.61 \\
\rowcolor{black!20}\textbf{0.5} & \textbf{13.5\% $\pm$ 0.71} & \textbf{95.4\% $\pm$ 0.36} & \textbf{96.5\% $\pm$ 0.44} \\
\bottomrule
\end{tabular}
\end{center}

\end{table}

As it can be seen, the most challenging experimental condition is the one which has the minimum time shift.~This is expected because it has increased the similarity between the genuine and impostor pairs, which has the inverse relation with the time shifted values used for impostor pair generation.

\subsection{Evaluation on the AVD dataset}

The proposed method was also evaluated on the \emph{AVD} dataset.~In all experiments, the training setup is the same as described in previous experiments.~For evaluation,~a 0.5-second time shift is used for generating the impostor pairs.~The years of 2014-2015 of the dataset are chosen for the experiments of this section.~In total 495 subjects are available in years of 2014-2015.~Among them, 201 subjects that are solely present in 2015 chosen to be as test subjects.~The rest of the subjects are used for creating the training data.

The setup for creating pairs is within-clip data generation,~e.g.,~the genuine and impostor pairs are built upon separated clips.~Each video clip and its corresponding audio only belong to one individual.~We deliberately regulate this setup for speaker-independent evaluation of audio-visual recognition.

Since the videos in the \emph{AVD} dataset are not scripted,~further data preprocessing is needed.~The data preprocessing includes Voice Activity Detection~(VAD) and elimination of the void sections of the visual stream in which no mouth area has been detected.~The two aforementioned preprocessing operations have been performed successively,~i.e.,~the data have been refined in two different phases.~The challenge was to maintain the corresponding audio-visual streams such that they have common timing characteristics.

\subsubsection*{Train and fine-tune on \emph{AVD} dataset}

\begin{figure}[t]
\begin{center}
   \includegraphics[width=1.0\linewidth]{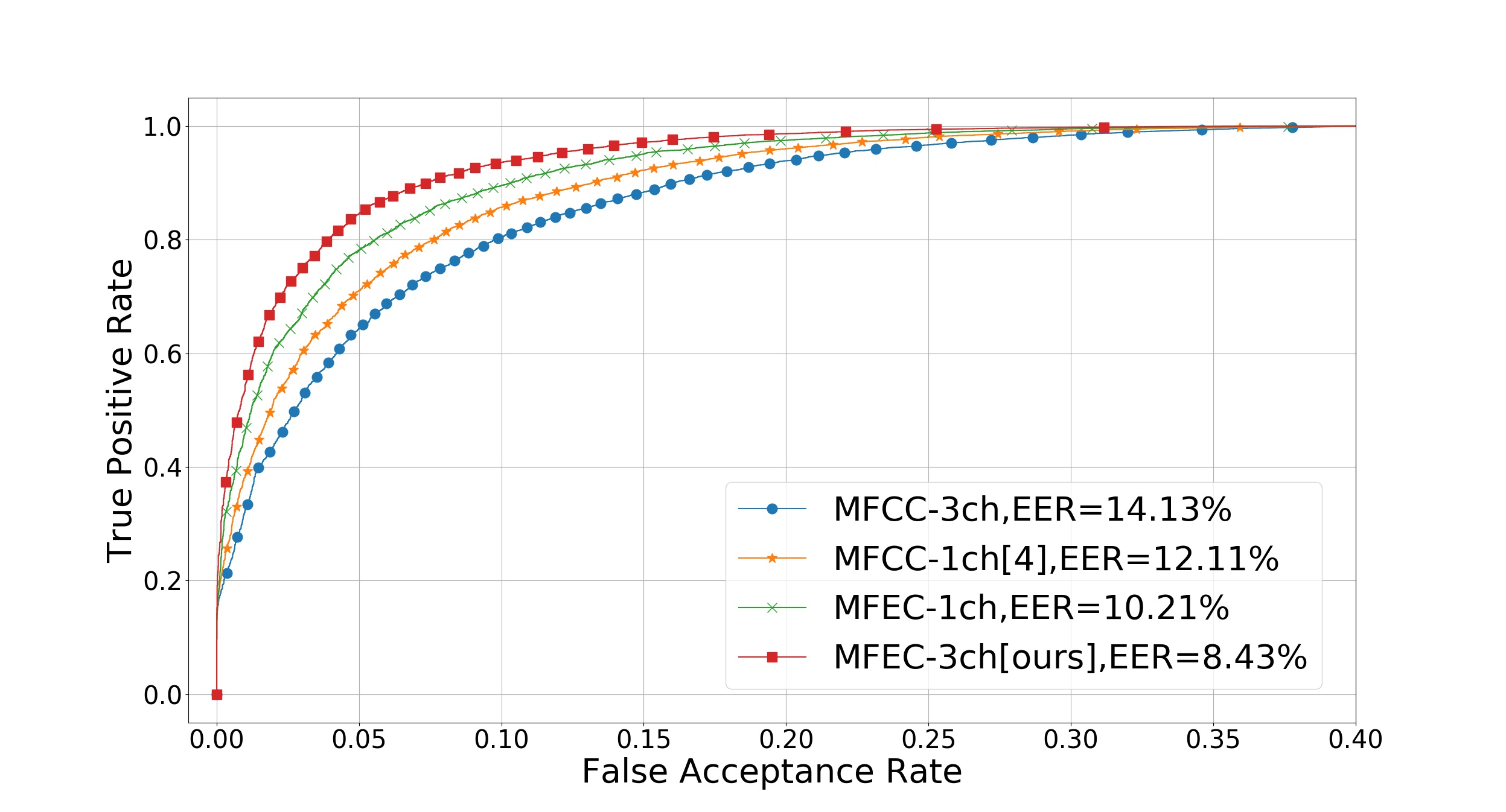}
\end{center}
   \caption{The ROC curve representation for fine-tuning on \emph{AVD} dataset.}
\label{fig:roc-fine-tuning}
\end{figure}

\begin{figure}[t]
\begin{center}
   \includegraphics[width=1.0\linewidth]{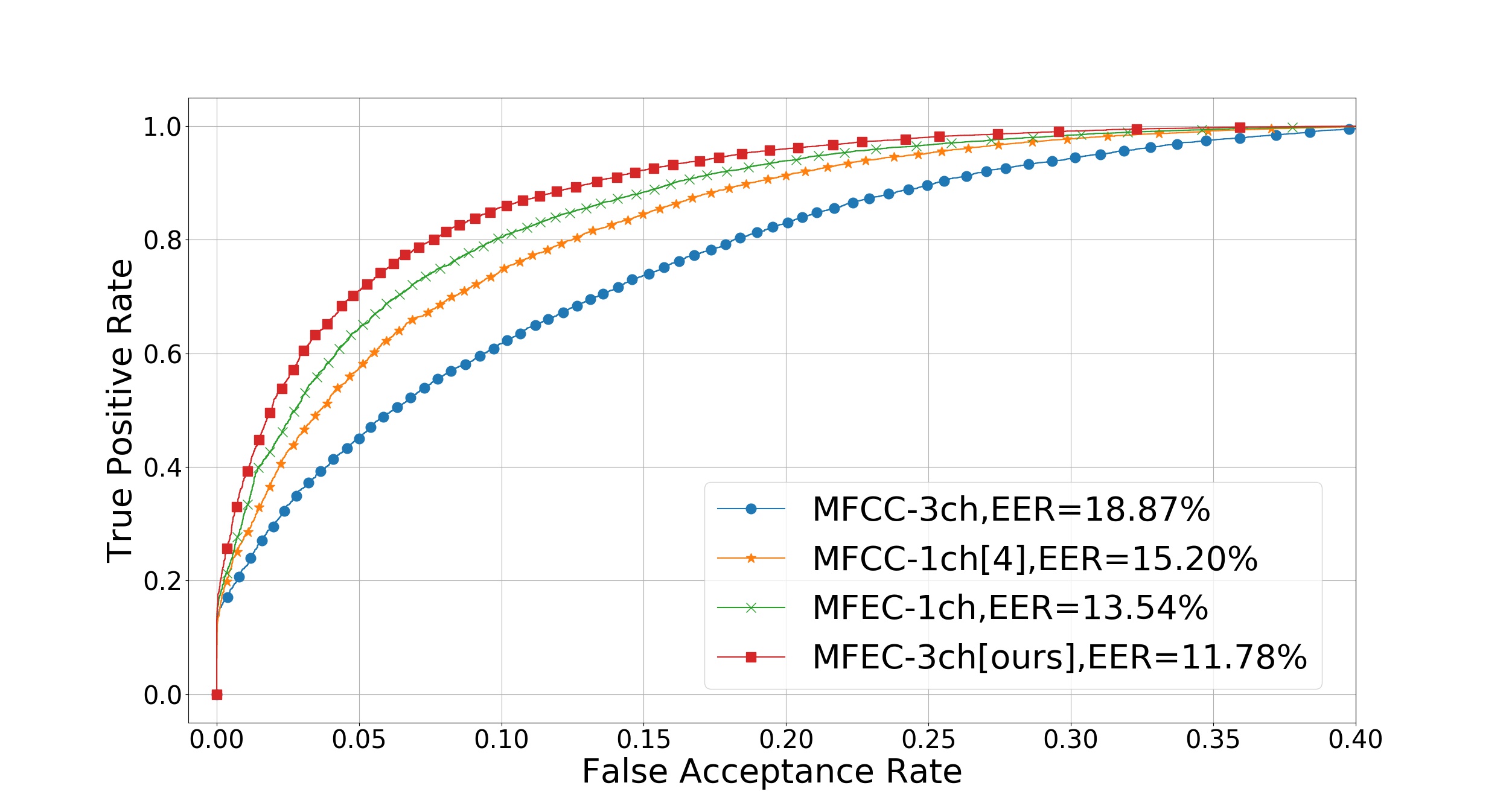}
\end{center}
   \caption{The ROC curve representation for training solely on \emph{AVD} dataset.}
\label{fig:roc-restricted}
\end{figure}


\begin{table}[h]
\caption[Table caption text]{Comparison of different methods on AVD dataset.}
\label{table:effect-of-feature-AVD}
\begin{center}
\addtolength{\tabcolsep}{-4pt}
\begin{tabular}{ccccc}
\toprule 
architecture & feature-depth & EER & AUC & AP \\
\hline
\midrule
\rowcolor{black!0} CCA\cite{sargin2007audiovisual} & MFCC-3 & 18.9\% $\pm$ 0.83 & 90.2\% $\pm$ 0.70 & 89.8\% $\pm$ 0.81 \\
\rowcolor{black!0} CoIA\cite{rua2009audio} & MFCC-3 & 18.3\% $\pm$ 0.69 & 90.4\% $\pm$ 0.78 & 90.1\% $\pm$ 0.64 \\
\rowcolor{black!5} CNN\cite{Chung16a} & MFCC-1 & 12.11\% $\pm$ 0.89 & 95.7\% $\pm$ 0.81 & 95.9\% $\pm$ 0.61 \\
\rowcolor{black!5} CNN\cite{Chung16a} & MFCC-3 & 14.13\% $\pm$ 1.03 & 94.7\% $\pm$ 1.12 & 95.0\% $\pm$ 1.26 \\
\rowcolor{black!10} 3D-CNN & MFEC-1 & 10.21\% $\pm$ 0.62 & 96.7\% $\pm$ 0.44 & 97.0\% $\pm$ 0.57 \\
\rowcolor{black!10} \textbf{3D-CNN} & \textbf{MFEC-3} & \textbf{8.43\% $\pm$ 0.66} & \textbf{98.0\% $\pm$ 0.41} & \textbf{98.5\% $\pm$ 0.51} \\
\bottomrule
\end{tabular}
\end{center}

\end{table}

In this section, we first evaluate our model on the \emph{AVD} dataset using the standard protocol of \emph{Restricted, No Outside Data}~\cite{LFWTechUpdate}.~This evaluation protocol is harsh since it assumes that no data from outside of the \emph{AVD} dataset will be used, and the use of feature extractors that have been trained on outside data is not allowed.~The results are demonstrated in Fig.~\ref{fig:roc-restricted}.

After experimenting using \emph{Restricted, No Outside Data} setup,~the training is done by fine-tuning the weights of the pre-trained network~(Section~\ref{Oxford Dataset Experiments}) by continuing the training on the \emph{AVD} dataset.~The results are depicted in Fig.~\ref{fig:roc-fine-tuning}.~For the fine-tuning, the learning rate has been set to $10^{-6}$ with no decay, and the training has been performed for 15 epochs of training data.~It is worth noting that using temporal derivatives for MFCC features downgraded the performance,~as observed in Fig.~\ref{fig:roc-fine-tuning}.~This can be related to local calculation of derivatives feature upon the non-local MFCC features. The results with comparison to other methods are summarized in Table~\ref{table:effect-of-feature-AVD}.


As can be observed in Table~\ref{table:effect-of-feature-AVD}, for the experiments on AVD dataset, the proposed method achieves relative improvements over 29\% on the Equal Error Rate~(EER) in comparison to the state-of-the-art method.

\section{Conclusion}

We have presented a novel coupled 3D convolutional architecture for audio-visual stream networks with convolutional fusion in temporal dimension~(by utilizing 3D convolutional and pooling operations) and coupling between the networks.~Experimental results on different data sets verified that the proposed architecture outperforms the other existing methods for audio-visual matching,~and moreover decreases the number of parameters significantly compared to the previously proposed methods.~Our performance results demonstrate the effectiveness of the joint learning of spatial and temporal information using 3D convolutions rather than naively combining them within the network.~The utilized local speech representative features are shown to be more promising for audio-visual recognition using convolutional neural networks. 



%

%

\section*{Acknowledgment}

This material is based upon work supported by the Center for Identification Technology Research and the National Science Foundation under Grant \#1650474.

\ifCLASSOPTIONcaptionsoff
  \newpage
\fi



%

%
%
{\bibliographystyle{IEEEtran}
\bibliography{bib}
}

\end{document}